\documentclass[runningheads,a4paper]{llncs}
\usepackage{multirow}
\usepackage[english]{babel}

\usepackage{amsmath}
\usepackage{amssymb}
\usepackage{graphicx}
\usepackage{multirow}
\usepackage{float}
\usepackage{caption}
\usepackage{hyperref}
\graphicspath{{imgs/}}

\begin{document}

\title{Ensembling Neural Networks for Digital Pathology Images Classification and Segmentation}
\titlerunning{Ensembling Neural Networks for Breast Cancer Histology}

\author{
	Gleb Makarchuk\thanks{Equal contribution}\inst{1, 3} \and 
    Vladimir Kondratenko\textsuperscript{*}\inst{1, 3} \and 
    Maxim Pisov\textsuperscript{*}\inst{1, 3} \and \\
    Artem Pimkin\textsuperscript{*}\inst{1, 3} \and 
    Egor Krivov \inst{1, 2, 3} \and 
    Mikhail Belyaev\inst{2, 1}
}

\authorrunning{
    G. Makarchuk, 
    V. Kondratenko,
    M. Pisov,
    A. Pimkin
    et al.
}

\institute{
Kharkevich Institute for Information Transmission Problems, Moscow, Russia \and
Skolkovo Institute of Science and Technology, Moscow, Russia \and 
Moscow Institute of Physics and Technology, Moscow, Russia
}
\maketitle

\begin{abstract}
In the last years, neural networks have proven to be a powerful framework for various image analysis problems. However, some application domains have specific limitations. Notably, digital pathology is an example of such fields due to tremendous image sizes and quite limited number of training examples available. In this paper, we adopt state-of-the-art convolutional neural networks (CNN) architectures for digital pathology images analysis. We propose to classify image patches to increase effective sample size and then to apply an ensembling technique to build prediction for the original images.  To validate the developed approaches, we conducted experiments with \textit{Breast Cancer Histology Challenge} dataset and obtained  90\% accuracy for the 4-class tissue classification task. 
\end{abstract}

\section{Introduction}
Histology is a key discipline in cancer diagnosis thanks to its ability to evaluate tissues anatomy. The classical approach involves glass slide microscopy and requires thoughtful analysis by a pathologist. Digital pathology imaging provides a statistically equivalent way to analyze tissues \cite{snead2016validation}, so it's a natural application area for machine learning methods \cite{madabhushi2016image}. Recent advances in deep learning suggest that these methods can be useful for a set of digital pathology image analysis problems including cell detection and counting and segmentation \cite{janowczyk2016deep}. One of the most important and challenging tasks is tissue classification, and recent studies (e.g., \cite{araujo2017classification}) demonstrate promising results. However, these image analysis problems differ from standard one in different ways. A crucial limiting factor is a combination of relatively small sample sizes (usually hundreds of examples) and extremely high resolution (the typical image size is 50000 x 50000). For comparison, the ImageNet dataset contains millions of 256 x 256 images \cite{deng2009imagenet}. This combination leads to high variance of deep learning models predictions and requires careful design of data processing pipelines. In this work, we propose two methods for digital pathology images segmentation and classification. Both methods include data preprocessing, intensive usage of modern deep learning architectures and an aggregation procedure for decreasing model variability. 
\section{Problem}
Generally, our task was to recognize benign and malignant formations on breast histology images. We were solving this problem in two formulations: image classification and segmentation.

\subsection{Data}
In this paper we use the   \textit{Breast Cancer Histology Challenge} (BACH-18) dataset which consists of \textit{hematoxylin and eosin} stained microscopy images as well as whole-slide images.
\label{subsec:data}
\subsubsection{Microscopy images}

The first subset consists of 400 microscopy images of shape $2048 \times 1536 \times 3$, where $3$ stands for the number of channels, the color space is RGB.

The images were obtained as patches from much larger microscopy images, similar to those described in subsection \ref{subsec:whole_slide}.

Every dataset entry is labeled as belonging to one of the four classes: Normal (n, 0), Benign (b, 1), Carcinoma in situ (is, 2) or Invasive Carcinoma (iv, 3). The labels are evenly distributed between the 400 images.

\begin{figure}
  \centering
  \includegraphics[width=0.95\textwidth]{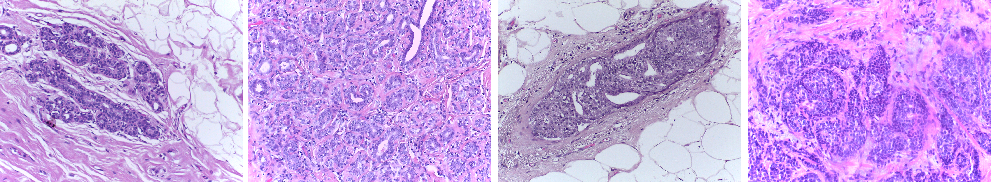}
  \caption{\label{fig:photo_sample}Samples from the first subset (from left to right): Normal, Benign, Carcinoma in situ and Invasive Carcinoma}
\end{figure}

\subsubsection{Whole-slide images}
\label{subsec:whole_slide}
The second subset consist of 20 whole-slide images of shape (approx.) $40000 \times 60000 \times 3$, where, similarly, $3$ stands for the number of channels, and the color space is RGB.

For half of the images segmentation masks of corresponding spatial shape are provided. Each pixel of the mask is labeled from 0 to 3 (from Normal to Invasive Carcinoma). During the model training only this half of the dataset was used.

\begin{figure}
  \centering
  \includegraphics[width=0.95\textwidth]{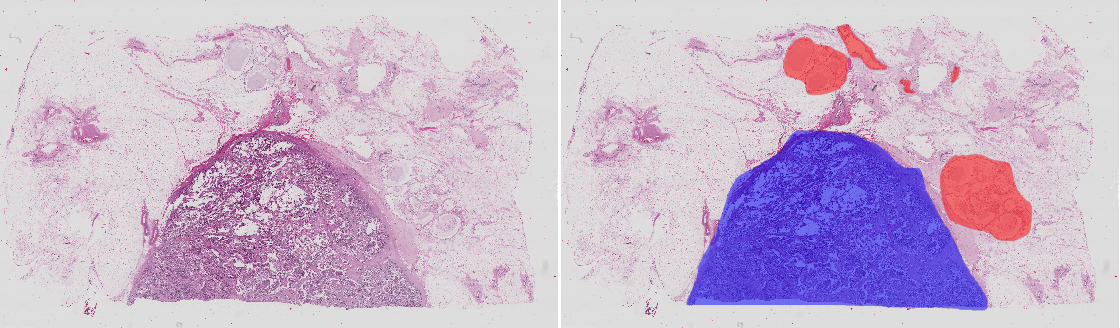}
  \caption{\label{fig:whole_slide}A sample from the second subset: whole-slide image (left) and the corresponding mask overlayed on the image (right). The blue color stands for Invasive Carcinoma, red - Benign, the rest - Normal. Note the roughness of the ground truth segmentation.}
\end{figure}

\section{Microscopy images classification}
\subsubsection{Preprocessing}
One of the common problems in work with histology slides is preprocessing. As it turned out, network pays attention to areas of inhomogeneity. For this reason, our image preparation was aimed at normalization and contrast enhancement of the slide. We tried a couple of well-known methods of digital pathology images preprocessing (e.g., \cite{preprocessing}), but it didn't increase the performance. Also, we tested simple data transformations like inversion, channel-wise mean subtraction, conversion to different color spaces, etc. A simple channel-wise mean subtraction provided the largest performance boost (about 10\% of accuracy score), so we ended up using only this preprocessing approach. 

\subsubsection{Training}
\label{subsec:training_photo}
We used two popular architectures for image classification: ResNet \cite{resnet} and DenseNet \cite{densenet}:

\textbf{ResNet} is a convolutional neural network which won the 1st place on ILSVRC 2015 classification task. We used ResNet34 implemented in \textit{torchvision} \footnote{http://pytorch.org/docs/0.3.0/torchvision/models.html} with slight architectural changes: we replaced the pooling layer before the fully-connected layer by an \textit{average spatial pyramid pooling layer} \cite{pyramid_pool}. We tried several levels of pyramid pooling depth from 1 (global average pooling) to 3.

\textbf{DenseNet} is a convolutional neural network which has a slightly smaller error rate on the ImageNet dataset than ResNet. We used DenseNet169 and DenseNet201 from \textit{torchvision}'s implementation with similar architectural changes as for ResNet.

In our experiments we used Adam optimizer \cite{adam} with an exponential-like learning rate policy: each 20 epochs the learning rate decreased by a factor of 2. Initially the learning rate was taken equal to 0.01.

Experiments have shown that feeding the images directly into the network yields poor results due to quite large images shape. In order to overcome this difficulty we chose to train out models on patches extracted from the original images: during  training each patch was randomly (with a 2D uniform distribution) extracted from an image, also picked at random, with the label being the same as for the original image. This approach led to a performance increase of about 15\%. 

Also, we observed that preprocessing each individual patch instead of preprocessing the whole image yields slightly better results.

Each network was trained on patches of shape $500\times500$ pixels, with batches of size $\approx10$ (this value differs between models). The training process lasted for 120 epochs, and every epoch 300 batches were fed into the network.


We also tried pretraining our models on similar datasets: BreakHis\footnote{https://omictools.com/breast-cancer-histopathological-database-tool}  
and \textit{Breast carcinoma histological images from the Department of Pathology}, "Agios Pavlos" General Hospital of Thessaloniki, Greece\cite{greece}. See Table \ref{tab:classification} for performance comparison.

\subsubsection{Model Selection and Stacking}
While building models, we experimented with different architectures, learning rate policies and pretraining. Thus, we ended up with 29 different models built and evaluated using 3-fold cross-validation (CV). We had a goal of combining these models in order to make more accurate predictions. 

During inference we deterministically extracted patches according to a grid with a stride of 100 pixels. Thus, every multiclass network would generate a matrix of shape $176\times4$ containing 4 class probabilities for 176 patches extracted from the image. Similarly, every one-vs-all network would generate a matrix of shape $176\times1$.

We extracted various features from these class probabilities predictions:
\begin{itemize}
    \item min, max and mean values of the probabilities of each class
    \item (for multiclass networks only) on how many patches each class has the highest probability
    \item 10, 25, 75 and 90 percentile probability values of each class
    \item on how many patches probabilities go above 15\% and 25\% threshold values for each class
\end{itemize}


After building the features, the problem was reduced to tabular data classification. So, for our final classifier, we have chosen XGBoost \cite{xgboost} as one of the state-of-the-art approaches for such tasks.

Our pipeline was the following:
\begin{itemize}
\item Choose reasonable XGBoost hyperparameters (based on cross-validation score) for the classifier built on top of all (29) models we have.
\item Use greedy search for model selection: keep removing models while the accuracy on CV keeps increasing.
\item Fine-tune the XGBoost hyperparameters on the remaining models set.
\end{itemize}

By following this procedure we reduced the number of models from 29 to 12. It is also worth mentioning that we compared accuracies for different sets of models and hyperparameters by averaging accuracy score from 10-fold CV across 20 different shuffles of the data to get statistically significant results (for such a small dataset) and thus optimize based on merit rather than on randomness. 

\subsubsection{Inference}
\label{subsubsec:final_pred}
Given the relatively small dataset, we decided that we might take advantage from retraining the networks on the whole dataset. However, this approach leaves no possibility to assess the stacking quality. 

As a trade-off, we decided to use 6-fold CV (instead of 3-fold CV), so that the network would see 83\% (vs 66\%) of data: a substantial increase in performance (compared to the 3-fold CV models) would mean that retraining on the whole dataset might be beneficial.

In the 6-fold CV setting, the performance of every individual network increased significantly. We also have fine-tuned the composition for 6-split networks that resulted in two more models being held out and slightly changed hyperparameters (see table \ref{tab:classification} for a comparison of networks' performances). However, the resulting ensemble classifier could not surpass the one build on top of 3-fold CV.

Nevertheless, for our final classifier we chose to average the patch predictions across all 6 networks and use the XGBoost classifier built on top of 6-fold CV. This approach is computationally inefficient, but allows us to reduce variance of the predictions.

\section{Whole-slide images segmentation}
\subsubsection{Preprocessing}
Similarly to the first problem, we use channel-wise mean subtraction as a preprocessing strategy. Also, given the unusually big images, we tried to downsample it by various factors: the downsampling by a factor of 40 along each spatial dimension proved to be very effective. Thereby, downsampled input is used in some ensemble models. 

\subsubsection{Training}
In our segmentation experiments we also used the same optimizer and learning rate policy as in section \ref{subsec:training_photo}.

In case when downsampling was included in the preprocessing pipeline the images were fed directly into the network, and the network was trained for 1500 epochs. Otherwise, the network was trained for 150 epochs with patches of shape $300\times300$ (40 patches per batch), similarly to the procedure described in section \ref{subsec:training_photo}.
\subsubsection{Models}
For the segmentation task, we introduce T-Net, a novel architecture based on U-Net \cite{unet}. It can be regarded as a generalization, which applies additional convolutions to the connections between the downsampling and upsampling branches.

\begin{figure}
  \centering
  \includegraphics[width=0.70\textwidth]{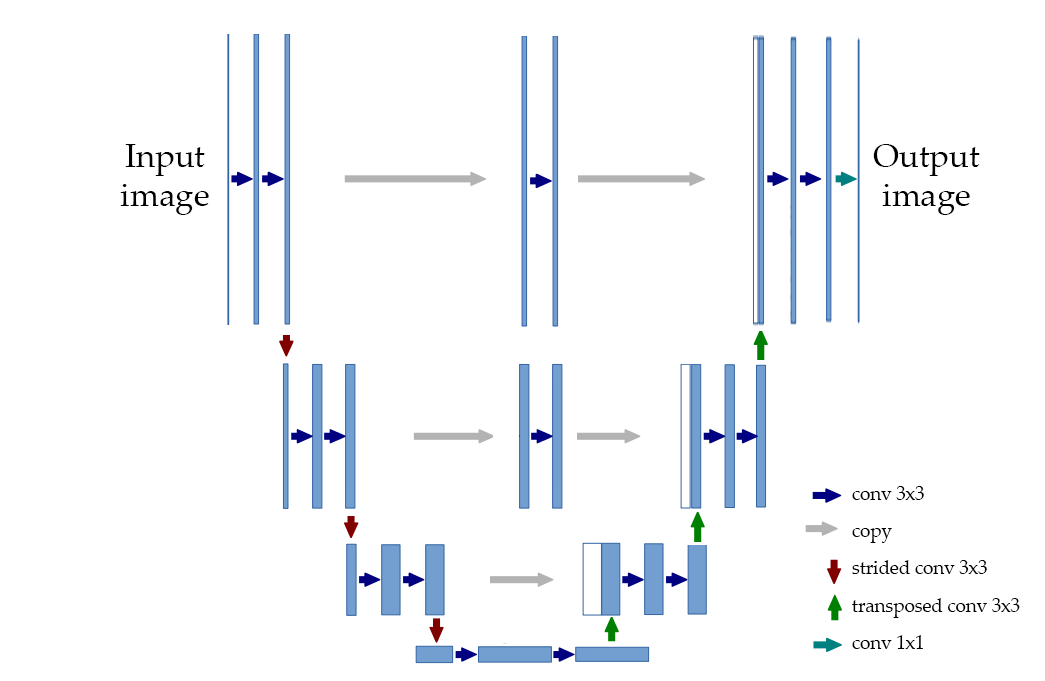}
  \caption{\label{fig:tnet_example}A schematic example of the T-Net architecture.}
\end{figure}


\label{subsubsec:wise_models}
\noindent We used 3 different models: 
\begin{itemize}
\item T-Net for binary segmentation Normal-vs-all trained on patches (T-Net 1).
\item T-Net for a similar task but trained on images downsampled by a factor of 40 along each dimension (T-Net 2). We also used a weighted-boundary log loss, which adds linearly decreasing weights to the pixels near the ground-truth regions' boundaries. 
Basically, it can be reduced to multiplying the ground truth mask by the corresponding weights and calculating the log loss for the resulting "ground truth".
\item T-Net for multiclass segmentation trained on images downsampled by a factor of 40 along each dimension (T-Net 3).
\end{itemize}

\subsubsection{Postprocessing}
\begin{figure} 
  \minipage{0.166\textwidth}
  \center{\includegraphics[width=\linewidth]{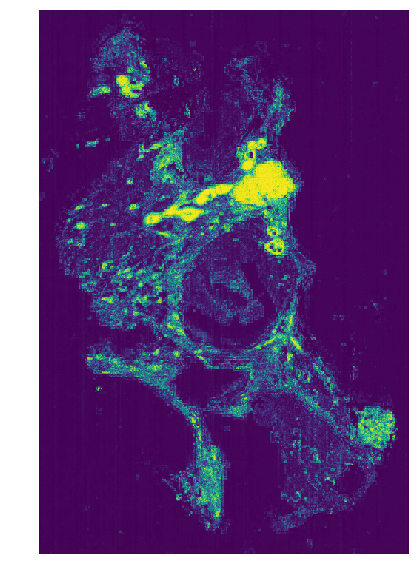}\\Original}
\endminipage\hfill
\minipage{0.166\textwidth}
  \center{\includegraphics[width=\linewidth]{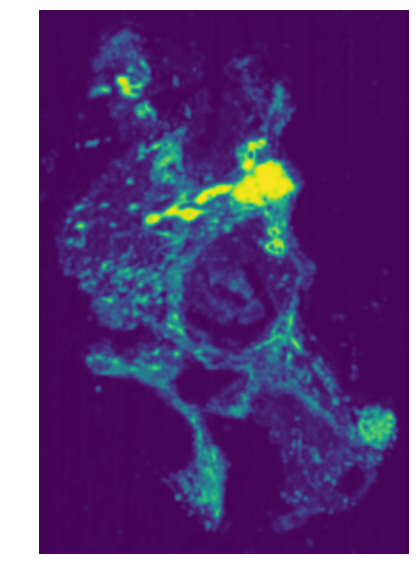}\\Blurred}
\endminipage\hfill
\minipage{0.166\textwidth}
  \center{\includegraphics[width=\linewidth]{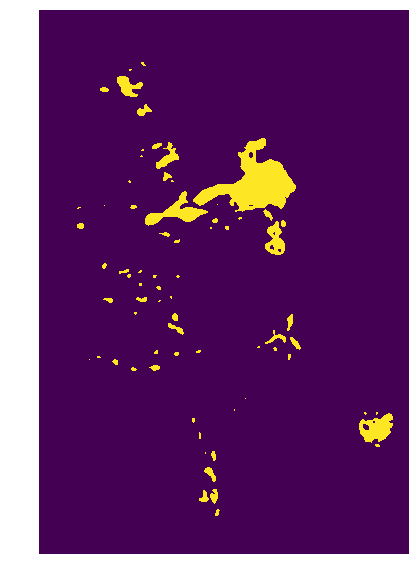}\\Thresholded}
\endminipage\hfill
\minipage{0.166\textwidth}
  \center{\includegraphics[width=\linewidth]{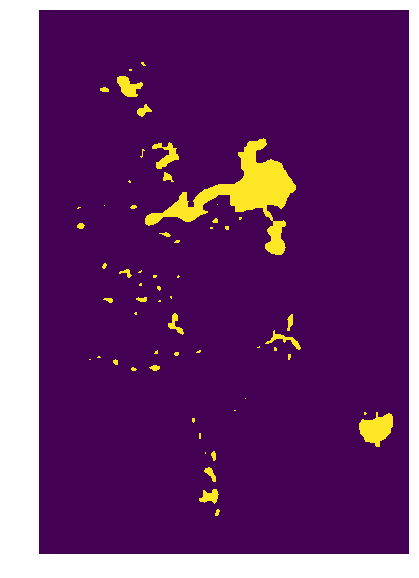}\\Closed}
\endminipage\hfill
\minipage{0.166\textwidth}
  \center{\includegraphics[width=\linewidth]{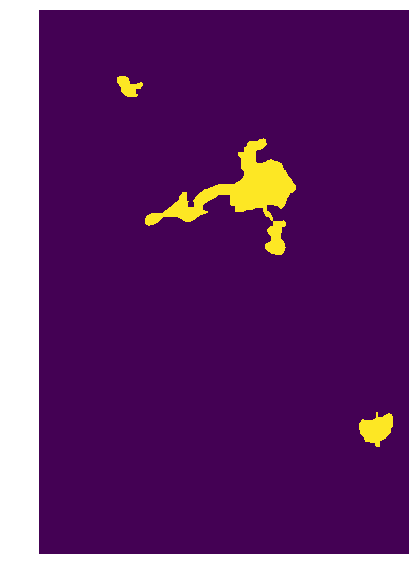}\\Result}
\endminipage\hfill
\minipage{0.166\textwidth}
  \center{\includegraphics[width=\linewidth]{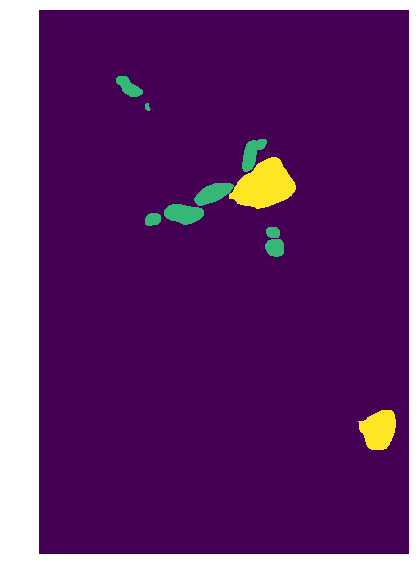}\\Ground Truth}
\endminipage\hfill
  \caption{\label{fig:postprocessing} Example of postprocessing. (Original) and (blurred) is the probability maps. Yellow on (Ground truth) is 'iv', green is 'is'. Yellow on the other images means abnormal.}
\end{figure}

While working with output of network trained on patches of the non-downsampled whole-slide images we faced the fact that output probability maps were too heterogeneous, which resulted in holes in segmented areas after thresholding, although ground truth consists of 1-connected domains. So, for primary processing we use Gaussian blur with square kernel of fixed size (processing hyperparameter) to reduce the hole sizes on the next steps. Then we threshold the probability map and get several clusters of areas with holes, many of which we merge with the morphological closing operation\cite{serra-closing-1983}. Finally, we discard the connected components with areas less than $\sqrt[a]{\overline{S^a}}$, where $\overline{S^a}$ is the mean of component areas in the power of $a$ ($a$ is a hyperparameter). The postprocessing steps are shown in Fig. \ref{fig:postprocessing} 

\subsubsection{Ensembling}
Due to strong class imbalance ($75\%-n:1\%-b:1\%-is:23\%-iv$) we focused our research on the Normal-vs-All task to adjust the output more precisely. Experiments have shown that the network trained on whole-slide images (T-Net 1) predicts quite "ragged" regions while the network trained on downsampled images (T-Net 2) predicts a lot of false-positive pixels.\\
\indent Our approach is T-Net ensemble that consists of blending these two models, letting them to compensate each other's mistakes, and transforming output binary mask into a multiclass one using (T-Net 3)'s prediction:
\begin{equation}
	Result = 3\cdot BinaryMask + (TNet3)\cdot (1 - BinaryMask)
\end{equation}
\indent However, given the fact that the metric proposed by the organizers of the BACH-18 challenge is biased towards the abnormal classes (1,2,3), we decided to use a much simpler approach - shifted blending, by setting the binary positive class to Invasive Carcinoma and the negative class to Benign:
\begin{equation}
	Result = 1 + 2\cdot BinaryMask
\end{equation}
It significantly increased the proposed metric. Also, since it is shifted we provide more common metrics for both approaches. See \ref{sub:results} for details.

Each model was trained and evaluated with 3-fold CV. To evaluate performance of the ensemble we used 5-fold CV (on top of the test predictions from 3-fold CV).

\section{Results}
\subsubsection{Microscopy images classification}
Table \ref{tab:classification} shows the models' accuracies. Note the significant performance boost gained from stacking.

\begin{table}[h]
\centering
\begin{tabular}{lc|c}
Network                                            & 3-fold CV & 6-fold CV  \\ \hline
ResNet 34                                          & .83     $\pm$ .01      & .86  $\pm$ .03                   \\
ResNet 34, pretrained on patches from whole slides & .83     $\pm$ .05      & .85  $\pm$ .04                   \\
ResNet 34, pretrained on BreakHis Dataset          & .81     $\pm$ .04      & .87  $\pm$ .06                   \\
ResNet 34, pretrained on Agios Pavlos dataset      & .83     $\pm$ .03      & .86  $\pm$ .03                   \\
ResNet 34, patch mean subtraction                  & .79     $\pm$ .04      & .83  $\pm$ .03                   \\
DenseNet 169                                       & .78     $\pm$ .06      & .85  $\pm$ .02                   \\
DenseNet 201                                       & .80     $\pm$ .03      & .82  $\pm$ .06                  \\ \hline
ResNet 34, Benign vs all   & .90    $\pm$ .01  & .91 $\pm$ .02                   \\
ResNet 34, InSitu vs all   & .90    $\pm$ .05  & .93 $\pm$ .02                   \\
ResNet 34, Invasive vs all & .90    $\pm$ .03  & .91 $\pm$ .03   \\ \hline
Ensemble (stacking) & \textbf{.90 $\pm$ .05}  & \textbf{.90 $\pm$ .05}
\end{tabular}
\caption{Models' accuracy for the microscopy images classification task: multiclass (first block), one-vs-all (second block), final ensemble built on top of 3- and 6-fold CV (third block) \label{tab:classification}}
\end{table}
\subsubsection{Whole-slide images segmentation}
\label{sub:results}
In the BACH-18 challenge the following metric is used:
\begin{equation}
BachScore = 1-\frac{\sum_i \big{|}pred_i-gt_i\big{|}}{\sum_i max\{gt_i, 3-gt_i\}\cdot \mathbb{I}[gt_i > 0 \text{ \& } pred_i > 0]},
\end{equation}
where the summation is performed across all the pixels and $gt_i, pred_i$ are the i-th pixel values of the ground truth and prediction respectively.

Table \ref{tab:segm} shows the models' performances according to BachScore, as well as the Dice score - a more common segmentation quality measure (it is computed for each channel separately and also from the point of "normal/abnormal" task).

\begin{table}[h]
\centering

\begin{tabular}{lccccc}
Network                                            & BachScore & Dice(b) & Dice(is) & Dice(iv) & Dice(abnormal) \\ \hline
T-Net 1                                          &  .54$\pm$.16                &       &   &.61$\pm$.19  & .69$\pm$.16            \\
T-Net 2                                           & .59$\pm$.20                   &       &   &.68$\pm$.21  &.71$\pm$.19           \\
T-Net 3                                          & .56$\pm$.19                    & .01$\pm$.01      & .03$\pm$.04  &.59$\pm$.21  &.66$\pm$.21            \\ \hline
T-Net ensemble                                     & .64$\pm$.20                   & .01$\pm$.02      & .01$\pm$.02  &.71$\pm$.23  &.72$\pm$.20           \\
T-Net shifted blending                              & \textbf{.70$\pm$.06}                   & \textbf{.04$\pm$.06}      & \textbf{.03$\pm$.04}  & \textbf{.73$\pm$.21}  & \textbf{.74$\pm$.18}           \\
\end{tabular}
\caption{\label{tab:segm}Segmentation results}
\end{table}

\section{Conclusion}




We proposed a two-stage procedure for \textbf{digital pathology images classification} problem. To increase effective sample size, we used random patches for training. The developed ensembling technique allowed us not only to increase the prediction quality due to averaging but also combine results for individual patches into the whole image prediction. In overall, a promising classification accuracy was obtained.


As for the \textbf{whole-slide images segmentation} task, we obtained controversial results: on the one hand we obtained promising Bach and Dice scores, but on the other hand most of our work was aimed at roughening the obtained predictions in accordance with the given labeling. Moreover, the results of our top performing ensemble were heavily improved by hard biasing which doesn't allow us to say how well this result depicts our method's performance


\bibliographystyle{splncs03}
\bibliography{main}

\end{document}